\documentclass[twoside,11pt]{article}

\usepackage{subfigure}
\usepackage{array}
\usepackage{booktabs}
\usepackage{enumitem}
\usepackage[abbrvbib, preprint]{ArXiv}
\usepackage[svgnames]{xcolor}

\usepackage{amsmath} 
\usepackage{natbib}
\usepackage{xcolor}
\usepackage{colortbl}
\usepackage{enumitem}
\usepackage{xcolor}
\usepackage{algpseudocode}

\DeclareFixedFont{\myfont}{OT1}{ptm}{m}{n}{8pt}
\DeclareFixedFont{\myfontb}{OT1}{ptm}{bx}{n}{8pt}

\usepackage{marvosym}
\makeatletter
\def\@fnsymbol#1{\ifcase#1\or \text{\Letter}\or *\or \dagger\or \ddagger\else\@arabic{#1}\fi}
\makeatother

\usepackage{lastpage}
\jmlrheading{23}{\ Preprint 2025}{1-\pageref{LastPage}}{1/21; Revised 5/22}{9/22}{21-0000}{}

\firstpageno{1}

\begin{document}

\title{CDEoH: Category-Driven Automatic Algorithm Design With Large Language Models}

\author{\name Yu-Nian Wang \email wangyunian@hhu.edu.cn \\
\addr Key Laboratory of Water Big Data Technology of Ministry of Water Resources\\
\addr College of Computer Science and Software Engineering, Hohai University, Nanjing, China
\AND
\name Shen-Huan Lyu\thanks{Corresponding author} \email lvsh@hhu.edu.cn \\
\addr Key Laboratory of Water Big Data Technology of Ministry of Water Resources\\
College of Computer Science and Software Engineering, Hohai University, Nanjing, China\\
\addr Department of Computer Science, City University of Hong Kong, Hong Kong, China\\
\addr State Key Laboratory for Novel Software Technology,
Nanjing University, Nanjing, China
\AND
\name Ning Chen \email che-n-ing@hhu.edu.cn \\
\addr Key Laboratory of Water Big Data Technology of Ministry of Water Resources\\
\addr College of Computer Science and Software Engineering, Hohai University, Nanjing, China
\AND
\name Jia-Le Xu \email xujl@hhu.edu.cn \\
\addr Key Laboratory of Water Big Data Technology of Ministry of Water Resources\\
\addr College of Computer Science and Software Engineering, Hohai University, Nanjing, China
\AND
\name Baoliu Ye \email yebl@nju.edu.cn \\
\addr State Key Laboratory for Novel Software Technology, Nanjing University, Nanjing, China
\AND
\name Qingfu Zhang \email qingfu.zhang@cityu.edu.hk \\
\addr Department of Computer Science, City University of Hong Kong, Hong Kong, China
}

\editor{My editor}

\maketitle

\begin{abstract}
With the rapid advancement of large language models (LLMs), LLM-based heuristic search methods have demonstrated strong capabilities in automated algorithm generation. However, their evolutionary processes often suffer from instability and premature convergence. Existing approaches mainly address this issue through prompt engineering or by jointly evolving thought and code, while largely overlooking the critical role of algorithmic category diversity in maintaining evolutionary stability. To this end, we propose Category-Driven Automatic Algorithm Design with Large Language Models (CDEoH), which explicitly models algorithm categories and jointly balances performance and category diversity in population management, enabling parallel exploration across multiple algorithmic paradigms. Extensive experiments on representative combinatorial optimization problems across multiple scales demonstrate that CDEoH effectively mitigates convergence toward a single evolutionary direction, significantly enhancing evolutionary stability and achieving consistently superior average performance across tasks and scales.
\end{abstract}


\section{Introduction}
Heuristic algorithms have long been central to solving complex optimization and search problems. Over the past decades, classical metaheuristics such as simulated annealing \citep{vanLaarhoven1987SA}, tabu search~\citep{glover1997tabusearch}, and iterated local search~\citep{lourenco2019ils} have achieved broad success in both industry and academia. However, different applications involve diverse constraints and objectives, necessitating manual design, adaptation, or tuning of heuristics for each task. This process relies heavily on expert knowledge, is time-consuming, and limits the practical deployment of heuristic methods.

To address this issue, Automatic Heuristic Design (AHD) aims to automatically select, tune, or construct heuristics for a given task class~\citep{stutzle2019automated}. Early methods include Genetic Programming (GP), a widely used technique~\citep{10.5555/138936,Guo2022HyperHeuristicSurvey}, which has been applied to job-shop scheduling~\citep{miyashita2000jobshop}, wind farm maintenance strategy design~\citep{ma2024gphyperheuristic}, hybrid flow-shop scheduling~\citep{liu2022automaticdesign}, and more. GP can evolve increasingly fit candidates via crossover and mutation, but its effectiveness is limited by issues such as program bloat and reliance on a predefined primitive set~\citep{luke2003modification,langdon2008gpintro}. A poorly designed primitive set can restrict the search space, while constructing a suitable one for complex problems remains challenging.

The emergence of large language models (LLMs) has opened new opportunities for automatic heuristic design (AHD). With strong semantic understanding and code-generation abilities~\citep{austin2021programsynthesislargelanguage,nijkamp2023codegenopenlargelanguage}, LLMs can express diverse heuristic strategies in both natural language and executable code, enabling more powerful AHD frameworks. Prior work has demonstrated the promise of LLM-based AHD~\citep{zhang2024evolutionarysearch}. Recent evolutionary paradigms include FunSearch, which uses a multi-island structure to generate and recombine programs and achieves breakthroughs on mathematical search problems~\citep{romera-paredes2024mathematical}, and EoH, which co-evolves “thoughts” and code with multiple prompting strategies to improve quality and reduce cost~\citep{pmlr-v235-liu24bs}. However, FunSearch incurs high computational overhead, while EoH’s single-island evolution can lead to limited search directions and premature convergence. AlphaEvolve extends FunSearch to full program evolution and achieves strong results on matrix multiplication search~\citep{novikov2025alphaevolve}. Despite these advances, existing methods generally lack explicit modeling and control of algorithmic diversity. HSEvo addresses this by introducing explicit diversity metrics, mapping algorithms into a vector space via an encoder and clustering them~\citep{dat2025hsevo}. In contrast, we feed algorithm thoughts and code into an LLM to directly classify algorithms using its semantic understanding, differing from HSEvo’s encoder-based approach.

In this paper, we emphasize the importance of algorithmic category diversity in evolutionary search. In practice, heuristic algorithms often fall into distinct categories (e.g., brute-force, greedy, and dynamic programming), each with different performance limits. Without considering such category structure, evolutionary search tends to converge prematurely to a single paradigm, limiting effective exploration of the heuristic space.

Motivated by this observation, we introduce algorithmic categories into LLM-based evolutionary design and propose Category-Driven Evolutionary Algorithm Design (CDEoH). CDEoH explicitly models and manages algorithm categories within a single, low-cost evolutionary population. For each candidate generated by the LLM, we induce its category and maintain a category pool to monitor coverage. During each generation, a category-based two-stage selection preserves high-performing algorithms while ensuring category-level diversity. We also introduce a lightweight reflection mechanism to repair or rewrite promising candidates that fail to execute due to LLM hallucinations.


Our main contributions are summarized as follows:
\begin{itemize}[itemsep=0pt, topsep=0pt, parsep=0pt] 
\item We propose CDEoH, a category-aware evolutionary framework for automatic heuristic design, which improves algorithmic diversity with low computational cost and enhances overall search quality.
\item We introduce simple yet effective strategies, including algorithm category induction, category-based population selection, and an LLM-based reflection mechanism, which can be readily integrated into existing evolutionary AHD methods.
\item We conduct extensive experiments on two AHD tasks at multiple scales, showing CDEoH consistently outperforms baselines, while ablations highlight category induction’s critical role in preserving diversity.
\end{itemize}

\section{Related Works}
\subsection{Traditional Hyper-Heuristics}
Traditional hyper-heuristic methods mainly fall into two categories: heuristic selection and heuristic generation. Heuristic selection chooses the best-performing heuristic from a predefined, human-designed set based on the current search state or historical performance~\citep{zhu2023surgical}, whereas heuristic generation constructs new heuristics by combining or evolving existing components under predefined rules~\citep{zhao2023survey}. Owing to their generality and cross-problem adaptability, hyper-heuristic frameworks have been widely applied across diverse optimization domains~\citep{de2025joint,de2025selection,luo2026hyper,zhao2025bayesian}. Nevertheless, their dependence on human-designed heuristics limits both the selection space and the potential for heuristic innovation.

\subsection{LLM-Based Hyper-Heuristics}
In recent years, large language models (LLMs) have made significant advances in semantic understanding and generation~\citep{naveed2025comprehensive}, accompanied by continuous improvements in their code generation capabilities~\citep{chen2023teaching,liventsev2023fully}. Prior studies have explored applying LLMs to code performance optimization and algorithmic problem solving in competitive programming~\citep{shypula2023learning,li2022competition,shinn2023reflexion}, and have further employed them as optimizers in search and decision-making processes~\citep{yang2023large}. In the context of algorithm selection, LLMs have also been used to identify suitable algorithms for specific problem instances via similarity~\citep{wu2023llm}. Moreover, LLMs have demonstrated strong performance in general task solving~\citep{yang2023intercode,zhang2023planning}. However, most existing approaches rely on single or static prompt engineering, making it difficult to systematically generate high-quality and consistently effective heuristic strategies.

\subsection{LLM-based Evolutionary Computation}
Evolutionary computation (EC) is a general optimization paradigm inspired by natural evolution~\citep{back1997handbook}, and recent studies show integrating EC with large language models (LLMs) substantially improves performance across diverse tasks, especially code generation~\citep{guo2023connecting,lehman2023evolution,hemberg2024evolving}. EC–LLM combinations have also been applied to practical problems, including strategy generation for control tasks~\citep{hu2025multimodal} and algorithm design for vehicle routing~\citep{li2025ars}. The most related work is EoH~\citep{pmlr-v235-liu24bs}, which introduces an LLM-driven evolutionary framework that decouples heuristic thoughts from code and uses multiple prompts to reduce optimization cost while maintaining strong performance. Building on EoH, CDEoH adds category induction and reflection mechanisms while retaining thought–code separation, yielding more diverse and stable heuristic exploration.

\section{Preliminaries}


We consider a specific task instance $\mathcal{T}$ to be solved by heuristic algorithms, such as an online bin packing problem where the number of items and bin capacity constraints are given. All feasible heuristic algorithms applicable to $\mathcal{T}$ constitute a search space, denoted by $\mathcal{H}$.

For any heuristic algorithm $h \in \mathcal{H}$, we evaluate its performance on the task instance $\mathcal{T}$ using an evaluation function $f$, which returns a scalar performance score $f(h)$. A larger value of $f(h)$ indicates better algorithmic performance.

Our objective is to automatically identify an optimal heuristic algorithm $h^*$ from the search space $\mathcal{H}$ such that its evaluation score on $\mathcal{T}$ is maximized. This objective can be formalized as the following optimization problem:
\[
h^* = \arg\max_{h \in \mathcal{H}} f(h).
\]

The above heuristic algorithm search problems typically exhibit extremely high computational complexity, and the task of finding optimal or near-optimal heuristic algorithms is generally regarded as NP-hard. As a result, traditional exact optimization methods are often impractical at realistic problem scales. In this context, evolutionary algorithms offer an effective search paradigm by balancing exploration and exploitation, enabling the exploration of complex, non-convex, and discrete search spaces without relying on gradient information or explicit problem structure, and gradually approaching high-performance heuristic algorithms. With the recent advances in large language models (LLMs), their strong semantic understanding and generation capabilities, when combined with evolutionary algorithms, further expand the exploration of the search space. Representative works include ReEvo~\citep{ye2024reevolargelanguagemodels}, FunSearch~\citep{romera-paredes2024mathematical}, and EoH~\citep{pmlr-v235-liu24bs}.

\section{Method}
\begin{figure*}[!htbp]
    \centering
    \includegraphics[width=0.8\textwidth]{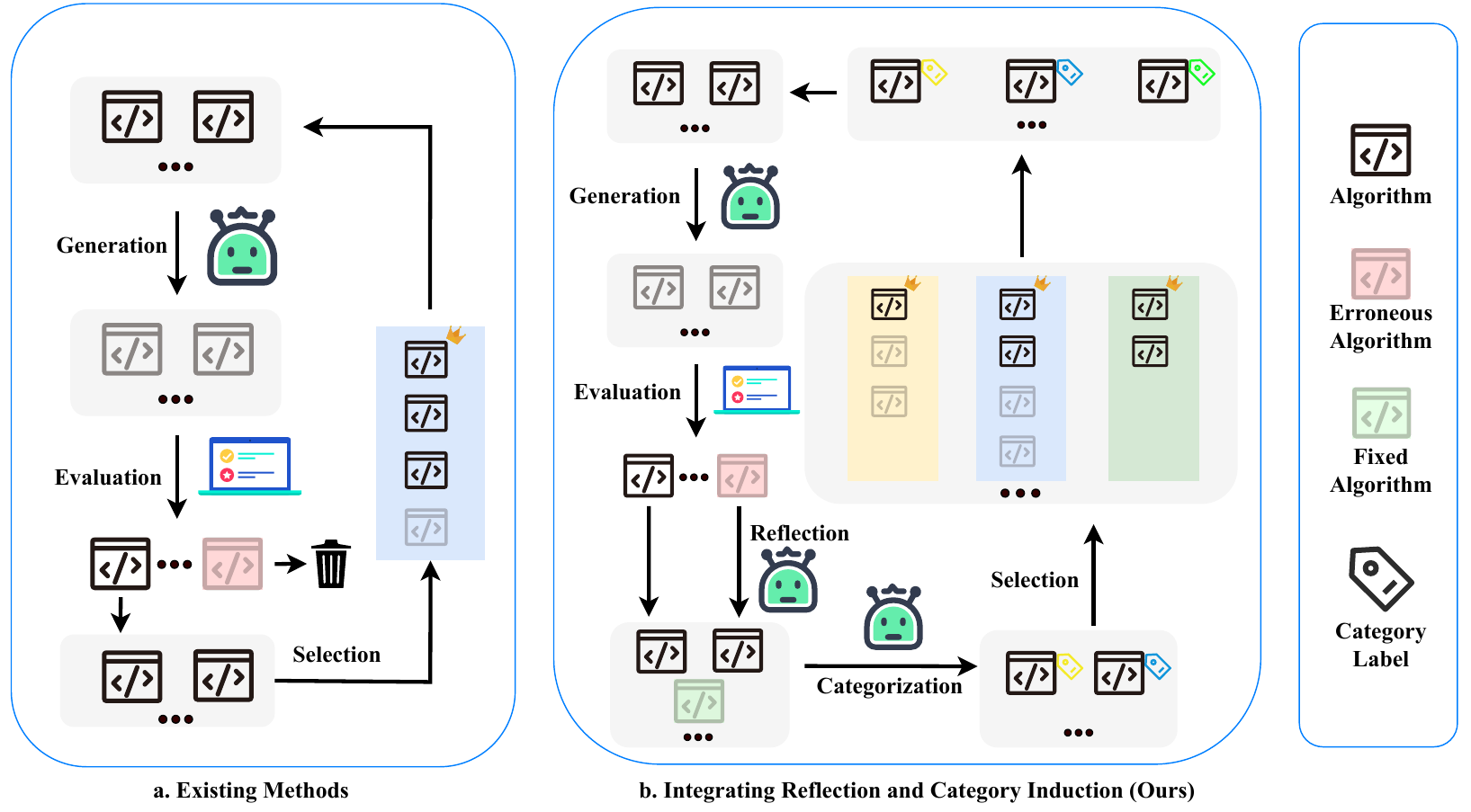}
    \caption{(a) Existing LLM-based automatic algorithm design methods adopt iterative search frameworks to optimize algorithms, where individuals in the population are not distinguished by category labels and erroneous algorithms are directly discarded. (b) CDEoH explicitly incorporates category labels to distinguish different algorithmic paradigms and employs a reflection mechanism to repair erroneous algorithms.}
    \label{fig:Method}
\end{figure*}

\subsection{Main Idea}
CDEoH enhances overall algorithmic diversity and search stability by explicitly categorizing algorithms and jointly considering algorithmic performance and category diversity during population selection. To achieve this goal, CDEoH incorporates the following key mechanisms:

\begin{itemize}
    \item Algorithm category modeling. 
    In addition to maintaining the thought and code of each algorithm, CDEoH explicitly records its algorithmic category. During each sampling step, CDEoH first follows the practice of EoH by invoking the LLM to generate a candidate algorithm, including its high-level design rationale (thought) and concrete implementation (code). Subsequently, CDEoH invokes the LLM again to perform category induction based on the algorithm's thought and code, and assigns a corresponding category label.
    \item Category-based two-stage selection strategy. 
    When evolving the population toward the next generation, CDEoH adopts a two-stage selection strategy to jointly balance performance and diversity. In the first stage, the top-performing (top-1) algorithm from each category is selected to ensure that high-quality representatives from different categories are preserved. In the second stage, the remaining algorithms are ranked according to a joint performance--diversity score, and a subset of high-scoring algorithms is selected to form the next-generation population.
    \item Reflection mechanism for error repair. 
    To prevent potentially valuable algorithms from being prematurely discarded due to LLM hallucinations or inappropriate modifications, CDEoH introduces a reflection mechanism for repairing erroneous algorithms. When an algorithm fails during evaluation, CDEoH submits the algorithm information together with the corresponding error messages to the LLM, guiding it to correct the errors so that the repaired algorithm may continue to participate in subsequent evolution.
\end{itemize}
Similar to EoH~\citep{pmlr-v235-liu24bs} and FunSearch~\citep{romera-paredes2024mathematical}, CDEoH integrates large language models into an evolutionary framework, leveraging their strong semantic understanding and code generation capabilities to continuously summarize, refine, and expand the population of algorithms.

Unlike existing methods that primarily rely on prompt engineering and stochastic generation to indirectly promote diversity, CDEoH explicitly maintains a category pool to manage all algorithm categories discovered during evolution. Category diversity is actively enforced through manually designed selection rules during population updates. This mechanism expands the search space at the level of algorithmic paradigms and enables CDEoH to more directly exploit category information compared to prior approaches such as FunSearch~\citep{romera-paredes2024mathematical} and EoH~\citep{pmlr-v235-liu24bs}. Fig.~\ref{fig:Method} illustrates the comparison between CDEoH and traditional methods.

\subsection{Overall Framework}

CDEoH maintains a population consisting of $N$ algorithmic individuals, denoted as
\[
P = \{h_1, h_2, \dots, h_N\}.
\]
Each algorithmic individual is evaluated on a set of task instances and assigned a fitness value $f(h_i)$, which measures its algorithmic performance.

During evolution, CDEoH samples parent algorithms from the current population to generate new candidate algorithms. Specifically, two types of prompts are employed for parent optimization: an improvement prompt and an innovation prompt. The improvement prompt focuses on local modifications of the parent algorithm, such as adjusting parameter configurations or refining existing structures. In contrast, the innovation prompt emphasizes restructuring the thought component of the algorithm, proposing new high-level problem-solving strategies and generating corresponding novel code implementations. We present the detailed prompts in Appendix.

Each newly generated algorithm that passes evaluation is subjected to category induction and assigned a category label. In each generation, CDEoH samples and evaluates at most $2N$ feasible new algorithms, added to the candidate pool. Then a category-based two-stage selection chooses $N$ algorithms to form the next population. This preserves high-performing algorithms while explicitly maintaining category diversity throughout evolution.

The evolutionary workflow of CDEoH is as follows:

\begin{enumerate}

    \item \textbf{Initialization.}
    CDEoH employs an initialization prompt that is repeatedly submitted to a large language model to obtain the initial population $P$, consisting of algorithms $h_1, h_2, \dots, h_N$.
    
    \item \textbf{Evolutionary Process.}
    For each generation, CDEoH executes the following steps:
    \begin{enumerate}[label=(\alph*)]
        \item \textbf{Algorithm Sampling.}
        For each algorithm in the current population, CDEoH performs:



        \begin{enumerate}[label=\roman*.]
            \item \textit{Innovation and Refinement Generation.}
            Given a population member, CDEoH invokes the LLM with an innovation-oriented prompt and a refinement-oriented prompt, generating two candidate algorithms.

            \item \textit{Evaluation and Reflection.}
            Each candidate is evaluated and assigned a performance score, which is stored as one of its attributes.
            If execution errors occur (e.g., due to LLM hallucinations or code inconsistencies), CDEoH employs a reflection prompt to submit the algorithm's core idea, code, and error message back to the LLM for attempted correction.
            This reflective repair process is repeated until the candidate executes successfully or a maximum reflection budget is reached.

            \item \textit{Category Induction.}
            Each new algorithm is assigned a category by the LLM and added to CDEoH's category pool.

            \item \textit{Population Augmentation.}
            The new algorithm, with its idea description, code, category, and score, is added to the population.
        \end{enumerate}

        \item \textbf{Population Management.}
        To ensure both high performance and sufficient diversity, CDEoH uses a category-driven two-stage selection strategy during each population iteration: first retaining the best algorithm from each category, then selecting the remaining candidates based on a joint performance-diversity score to form the next-generation population.
    \end{enumerate}

    \item \textbf{Termination.}
    The evolution terminates when the number of samples or the number of generations in the population reaches the predefined maximum.
\end{enumerate}

\subsection{Category Management}

Premature convergence and susceptibility to local optima have long been recognized as core challenges in evolutionary algorithms. These issues primarily stem from the rapid loss of population diversity, which severely restricts effective exploration of the search space. Existing approaches attempt to mitigate this problem through structural or prompt-level designs. For example, FunSearch adopts multi-island parallel evolution to maintain solution dispersion~\citep{romera-paredes2024mathematical}, while EoH employs improvement-oriented and innovation-oriented prompts to guide the model toward generating diverse algorithms~\citep{pmlr-v235-liu24bs}. Nevertheless, such methods still rely largely on implicit mechanisms to preserve diversity, making it difficult to systematically cover distinct algorithmic categories. As a result, algorithmic homogenization or unstable generation quality may still emerge during long-term evolution.

To explicitly model algorithmic diversity, we observe that most heuristic algorithms can be clearly categorized according to their core ideas and code structures, such as greedy methods, brute-force search, or dynamic programming. Moreover, different categories often correspond to markedly different performance ceilings. Motivated by this observation, CDEoH assigns each algorithmic individual a discrete category label that characterizes its underlying algorithmic paradigm.

Formally, let an algorithmic individual $h_i$ be represented by its high-level conceptual description $\text{thought}_i$ and its concrete code implementation $\text{code}_i$. Its category is then defined as
\[
c_i = \mathcal{C}(\text{thought}_i, \text{code}_i),
\]
where $\mathcal{C}(\cdot)$ denotes an algorithm classification mapping implemented by a large language model.

Based on this category definition, CDEoH maintains a category pool throughout the evolutionary process to record all algorithm categories discovered so far. This pool is dynamically expanded as evolution proceeds and serves as an important basis for subsequent population selection, thereby explicitly constraining category coverage.

Specifically, let the set of categories discovered up to generation $t$ be
\[
\mathcal{K}^{(t)} = \{ c_i \mid h_i \in P^{(t)} \}.
\]
When a newly generated algorithm $h_{\text{new}}$ completes evaluation, if its category $c_{\text{new}} \notin \mathcal{K}^{(t)}$, CDEoH adds this category to the category pool and treats it as a new algorithmic paradigm. In this way, the structural diversity of the search space is continuously expanded.

\begin{figure*}[!t]
    \centering
    \includegraphics[width=1.0\textwidth]{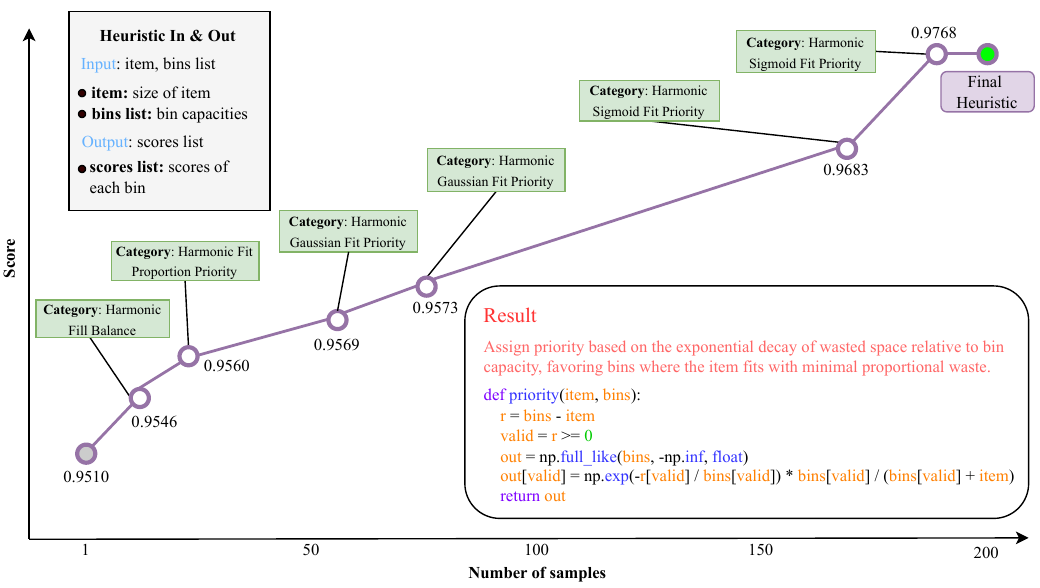}
    \caption{One evolutionary run of CDEoH on the online bin packing problem. We list the algorithm categories and present the detailed thoughts and code of the best-performing algorithm.}
    \label{fig:Line_Chart}
\end{figure*}

\subsection{Population Management}

During the transition of the population from generation $t$ to generation $t{+}1$, we aim to account for both algorithmic performance and category diversity in the selection process. On the one hand, selection strategies solely based on performance tend to drive the population to rapidly converge to a single algorithmic paradigm, thereby increasing the risk of being trapped in local optima. On the other hand, explicitly encouraging the preservation of diverse algorithm categories facilitates sustained exploration along multiple directions of the search space, fundamentally alleviating premature convergence. Motivated by these considerations, CDEoH introduces a category-driven two-stage selection strategy that explicitly maintains algorithmic category diversity while ensuring high performance.

\textbf{I. Category-wise Elitism.}
Different algorithm categories often correspond to distinct modeling assumptions and performance ceilings. During the early or intermediate stages of evolution, even if a category does not yet exhibit competitive overall performance, its best-performing individual may still possess substantial potential for further improvement in subsequent generations. Consequently, directly discarding all algorithms from a given category during selection may prematurely shrink the search space and lead to irreversible information loss. To address this issue, CDEoH incorporates a category-level elitism mechanism in the first selection stage, ensuring that each discovered algorithm category retains at least its current best representative.

Let $\mathcal{H}$ denote the current candidate algorithm set, and let $\mathcal{K}$ be the set of algorithm categories. For each category $c \in \mathcal{K}$, the best algorithm in that category is defined as
\[
h_c^{*} = \operatorname*{arg\,max}_{h_i \in \mathcal{H}, C(i)=c} f(h_i)
\]
where $f(h_i)$ denotes the performance score of algorithm $h_i$, and $C(i)$ indicates its associated category.

This yields the category-elite set
\[
\mathcal{E} \;=\; \{ h_c^{*} \mid c \in \mathcal{K} \}.
\]
CDEoH then ranks the algorithms in $\mathcal{E}$ according to their original performance scores $f(h_i)$ and directly retains the top $k$ algorithms as part of the next-generation population, where $N \ge k$ and $N$ denotes the population size.

By adding category constraints to selection, this stage preserves high-quality individuals from different paradigms, enforcing structural diversity and preventing premature convergence to one category.

\textbf{II. Performance--Diversity Selection.}
After completing category-wise elitism, it is still necessary to select additional individuals from the remaining candidates to fill the next-generation population to a fixed size. Purely performance-based ranking may again induce category concentration, whereas overly emphasizing diversity may introduce a large number of low-quality algorithms. To balance these factors, CDEoH designs a performance--diversity joint scoring mechanism that enables a controllable trade-off between the two objectives.

For each algorithm $h_i$ not selected in the first stage, CDEoH computes a joint score
\[
S_i \;=\; \frac{f_i - f_{\min}}{f_{\max} - f_{\min}} \;+\; \lambda \cdot \frac{1}{\lvert C(i) \rvert},
\]
where $f_{\max}$ and $f_{\min}$ denote the maximum and minimum raw performance scores among the current candidate algorithms, respectively; $\lvert C(i) \rvert$ is the number of algorithms belonging to the same category as $h_i$; and $\lambda$ is a hyperparameter that balances performance and diversity.

The scoring function consists of two components. The first term normalizes algorithm performance to mitigate scale differences across tasks or evaluation settings. The second term explicitly favors algorithms from rare or newly discovered categories, increasing their probability of being retained and thereby expanding exploration into underrepresented regions of the search space. When $\lambda = 0$, the strategy degenerates into a purely performance-driven selection scheme, whose behavior closely resembles the selection mechanism used in EoH~\citep{pmlr-v235-liu24bs}.

Finally, CDEoH ranks the remaining algorithms according to their joint scores $S_i$ and selects the top $N - k$ algorithms to complete the next-generation population.

\subsection{Reflection Mechanism}

During the evolutionary process, newly generated algorithmic individuals may suffer from execution failures or logical flaws due to hallucinations of large language models or inappropriate modification attempts. Existing evolutionary frameworks typically discard such erroneous algorithms to prevent interference with subsequent evolution. However, this strategy may prematurely eliminate candidates that are promising at the level of algorithmic ideas but flawed only in implementation details, thereby reducing search efficiency.

To address this issue, CDEoH introduces a reflection mechanism for automatic repair of erroneous algorithms during evolution. The core idea is to treat algorithm repair as a conditional generation process driven by a large language model, which performs targeted modifications to the original algorithm conditioned on the observed error context.

Specifically, when an algorithmic individual $h$ fails during evaluation and produces an error message $e$, CDEoH constructs a reflection input that includes the algorithm's high-level description (thought), its code implementation (code), and the corresponding error information. The large language model is then prompted to generate a repaired algorithmic individual. This process can be formalized as
\[
h' \;\sim\; \mathcal{R}_{\mathrm{LLM}}(h,\, e),
\]
where $\mathcal{R}_{\mathrm{LLM}}(\cdot)$ denotes a reflection operator parameterized by the LLM, which revises algorithm $h$ under error $e$ to produce a new candidate $h'$.

If the repaired algorithm still fails to pass evaluation, CDEoH repeats the reflection process within a predefined reflection budget $B$, until the algorithm executes successfully or the maximum number of reflection attempts is reached. By incorporating this reflection mechanism, CDEoH is able to recover and refine high-potential but imperfectly implemented algorithms with low additional computational overhead, thereby improving the effectiveness and stability of the overall evolutionary process.

\section{Experiments}
\subsection{Tasks and Instances}
We conducted systematic experiments on the Online Bin Packing (OBP) problem and the Traveling Salesman Problem (TSP). For each problem, the evolutionary process was run independently 10 times under different experimental settings, and the best performance was reported. The problems and evaluation protocols are detailed as follows.

\begin{itemize}
    \item \textbf{OBP.} In OBP, items arrive sequentially and must be assigned online to fixed-capacity bins, aiming to minimize the total number of bins used. Heuristic algorithms select a bin for each item. Performance is evaluated by the relative gap to a theoretical lower bound on the optimal number of bins, computed following Martello and Toth~\citep{martello1990lowerbounds}. The benchmark includes five instances generated from the Weibull distribution~\citep{romera-paredes2024mathematical}, with bin capacities of 100 and 500 and item counts from 1,000 to 10,000. The evaluation protocol follows~\citet{pmlr-v235-liu24bs}. We provide the task description and template code for OBP in Appendix.

    \item \textbf{TSP.} TSP aims to find the shortest possible closed tour visiting each city exactly once. We adopt constructive heuristics from prior work, which build solutions incrementally by selecting the next city to visit at each step. Performance is measured by the average relative gap to reference solutions produced by the LKH solver~\citep{helsgaun2017lkh_extension}. The benchmark includes sixteen instances with city counts ranging from 50 to 500. City coordinates are sampled from $[0,1]$, and Gaussian instances are generated following~\citet{bi2023learninggeneralizablemodelsvehicle}. Detailed task descriptions and template code for TSP are provided in Appendix.
\end{itemize}

\begin{table}[!t]
\centering
\small
\caption{OBP results under different problem settings. The relative gap between the number of bins used by different heuristics and the lower bound on instances generated from a Weibull distribution (lower is better). In each column, the best results are highlighted in bold, and the second-best results are shaded.}
\resizebox{\columnwidth}{!}{
\begin{tabular}{lcccccc}
\toprule
Method & 1kC100 & 1kC500 & 5kC100 & 5kC500 & 10kC100 & 10kC500 \\
\midrule
EoH 
& \cellcolor{gray!20}2.476 
& \cellcolor{gray!20}0.991 
& 2.406 
& 0.497 
& 1.071 
& 0.448 \\
FunSearch 
& 3.516 
& \cellcolor{gray!20}0.991 
& 3.878 
& \cellcolor{gray!20}0.447 
& 1.820 
& \cellcolor{gray!20}0.423 \\
ReEvo 
& 3.171 
& \cellcolor{gray!20}0.991 
& \cellcolor{gray!20}1.611 
& 0.497 
& \textbf{0.478} 
& \textbf{0.075} \\
CDEoH 
& \textbf{2.378} 
& \textbf{0.744} 
& \textbf{1.054} 
& \textbf{0.099} 
& \cellcolor{gray!20}0.483 
& \textbf{0.075} \\
\bottomrule
\end{tabular}
}
\label{tab:obp_results}
\end{table}

\begin{table}[!t]
\centering
\small
\caption{TSP results under different problem settings. The table compares the relative distance between the tour lengths generated by each heuristic and those produced by the LKH algorithm (lower is better). In each column, the best results are highlighted in bold, and the second-best results are shaded.}
\begin{tabular}{lcccc}
\toprule
Method & size50 & size100 & size200 & size500 \\
\midrule
EoH 
& \cellcolor{gray!20}10.060 
& \cellcolor{gray!20}13.097 
& \textbf{15.640} 
& \cellcolor{gray!20}21.448 \\
FunSearch 
& 14.921 
& 22.796 
& 17.680 
& 22.513 \\
ReEvo 
& 12.053 
& 15.249 
& 17.024 
& 22.248 \\
CDEoH 
& \textbf{9.226} 
& \textbf{12.328} 
& \cellcolor{gray!20}15.922 
& \textbf{17.134} \\
\bottomrule
\end{tabular}
\label{tab:tsp_results}
\end{table}

\subsection{Compared Methods and Settings}
We compared several recent representative LLM-based automatic heuristic design (AHD) methods, including ReEvo~\citep{ye2024reevolargelanguagemodels}, FunSearch~\citep{romera-paredes2024mathematical}, and EoH~\citep{pmlr-v235-liu24bs}.

All experiments were conducted on the LLM4AD platform, using DeepSeek-Chat as the underlying language model for all methods. The maximum sampling budget was set to 200 for each method (with minor differences). Except for FunSearch, all methods used a fixed population size of 10; FunSearch’s population size was dynamically adjusted via its multi-island mechanism. During CDEoH evolution, the top-1 algorithms from the four best categories were preserved, and $\lambda$ was set to 0.7. Each method was independently run 10 times under different problem scales, and the average performance was reported.

It is worth noting that CDEoH does not rely on crossover operations for sampling algorithm individuals within the population, and algorithms are independent of each other, which enables high parallelism. Therefore, when the computational cost of evaluation is relatively low, CDEoH can significantly reduce the overall evolution time. Fig.~\ref{fig:Line_Chart} illustrates an example of the evolutionary process of CDEoH.

\begin{table}[!t]
\centering
\small
\caption{Results of the ablation study on the online bin packing problem. The table reports comparisons of CDEoH with respect to the category mechanism and the reflection mechanism.}
\resizebox{\columnwidth}{!}{
\begin{tabular}{lcccccc}
\toprule
Method & 10kC100 & 10kC500 & 1kC100 & 1kC500 & 5kC100 & 5kC500 \\
\midrule
nocategory & 0.458 & \textbf{0.075} & 3.019 & 0.991 & 1.083 & 0.248 \\
noreflection & \textbf{0.368} & \textbf{0.075} & 2.969 & 0.991 & 1.163 & 0.248 \\
CDEoH & 0.483 & \textbf{0.075} & \textbf{2.378} & \textbf{0.744} & \textbf{1.054} & \textbf{0.099} \\
\bottomrule
\end{tabular}
}
\label{tab:ablation_results}
\end{table}

\subsection{Results}

Table~\ref{tab:obp_results} presents the experimental results for the online bin packing problem under different item scales and bin capacities. It can be observed that CDEoH demonstrates stable performance advantages on the OBP task. In almost all evaluated settings, CDEoH achieves strong results. Under multiple problem parameter configurations, CDEoH substantially reduces the relative gap compared with EoH and FunSearch, indicating strong robustness in high-dimensional online decision-making scenarios. Under the 10kC100 setting, CDEoH also achieves performance comparable to the best-performing method. These results suggest that CDEoH is particularly well suited for online combinatorial optimization problems, where the joint modeling of algorithm diversity and performance during evolution helps avoid premature convergence and improve solution quality.

Table~\ref{tab:tsp_results} reports the experimental results for the TSP under different problem sizes. CDEoH similarly achieves strong performance across all tested scales. In the size50, size100, and size500 settings, CDEoH attains consistently competitive results. Notably, under the large-scale size500 setting, the advantage of CDEoH becomes more pronounced, indicating that the method maintains stable performance as problem scale increases. In the size200 setting, where CDEoH ranks second, its performance remains close to that of the best method. These results further demonstrate that CDEoH achieves a good balance between heuristic search space exploration and solution quality, exhibiting strong generalization ability and scalability.

\subsection{Ablation Study}
We conduct ablation experiments on OBP to investigate the contribution of the category-driven mechanism and the reflection module in CDEoH under different problem scales. Two ablated variants are considered: nocategory, which removes category-based guidance, and noreflection, which disables the reflection process during heuristic evolution. The full CDEoH integrates both components.

Table~\ref{tab:ablation_results} reports the gap between the best results and the lower bound under six OBP settings. The results reveal that the impact of different components varies with problem scale. On large-scale instances (10kC100 and 10kC500), the ablated variants achieve comparable or slightly better performance than the full model, suggesting that the additional structural guidance introduced by category modeling and reflection may impose extra constraints when the search space is extremely large.

In contrast, as the problem scale decreases or the constraint becomes tighter (e.g., 1kC100, 1kC500, and 5kC500), CDEoH consistently outperforms both ablated variants, achieving significantly smaller gaps to the lower bound. This indicates that category-aware guidance and reflection are particularly effective in structured or moderately sized settings, where they help refine heuristic behaviors and improve solution quality.

Notably, the reflection module plays a critical role in more constrained settings, as evidenced by the substantial performance gap between noreflection and the full model on 1kC500 and 5kC500. Overall, these results suggest that the proposed components contribute complementary benefits, improving robustness and solution quality across a wide range of problem configurations, while exhibiting different trade-offs at extreme scales.

\section{Discussion and Future Work}
This work explicitly incorporates algorithm categories into an LLM-driven heuristic evolution process and jointly considers performance and category diversity in population management, thereby improving search stability and alleviating premature convergence. The proposed method is simple and easy to use, can be integrated into existing approaches, and yields performance improvements under most experimental settings. By exploring multiple category paradigms in parallel during evolution, CDEoH avoids local optima caused by reliance on a single paradigm.

However, experimental results show that CDEoH does not consistently achieve the best performance across all tasks and parameter settings, indicating room for improvement in expanding the breadth of paradigm exploration. In addition, ablation studies reveal that the proposed mechanisms may impose performance limitations in a small number of cases, suggesting the need for further refinement. Future work may focus on refining the proposed mechanisms, exploring finer-grained or hierarchical category modeling, and investigating category induction methods that do not rely on LLMs to mitigate potential hallucination issues.

\section{Conclusion}
This work proposes CDEoH, a method that performs category induction for evolving algorithms within the EoH framework and jointly considers performance and category diversity during population iteration. Compared with prior evolutionary approaches that ignore algorithm category information, CDEoH effectively mitigates premature convergence and local optima during evolution. Extensive experiments on combinatorial optimization problems demonstrate that CDEoH achieves competitive performance in most experimental settings and attains the best results in multiple scenarios. This study further highlights the potential value of incorporating algorithm category induction into LLM-based evolutionary methods.



\bibliography{main}

\end{document}